\documentclass[reprint,amsmath,amssymb,aps]{revtex4-2}

\usepackage{graphicx}
\usepackage{dcolumn}
\usepackage{bm}
\usepackage{hyperref}
\usepackage{lipsum}
\usepackage{subcaption}  
\usepackage{caption}         
\usepackage{float}
\usepackage{tikz}
\usepackage{ragged2e}
\usetikzlibrary{positioning, arrows.meta}  
\usepackage{relsize}  
\usepackage{overpic}  
\usepackage{placeins}

\begin{document}

\preprint{APS/123-QED}

\title{Different Paths, Same Destination: Designing New Physics-Inspired Dynamical Systems with Engineered Stability to Minimize the Ising Hamiltonian}

\author{E.M.H.E.B. Ekanayake}
\affiliation{%
University of Virginia, Charlottesville, VA, USA
}%

\author{N. Shukla}
\affiliation{%
University of Virginia, Charlottesville, VA, USA
}%

\begin{abstract}

Oscillator Ising machines (OIMs) represent an exemplar case of using physics-inspired non-linear dynamical systems to solve computationally challenging combinatorial optimization problems (COPs). The computational performance of such systems is highly sensitive to the underlying dynamical properties, the topology of the input graph, and their relative compatibility. In this work, we explore the concept of designing different dynamical systems that minimize the same objective function but exhibit drastically different dynamical properties. Our goal is to leverage this diversification in dynamics to reduce the sensitivity of the computational performance to the underlying graph, and subsequently, enhance the overall effectiveness of such physics-based computational methods. To this end, we introduce a novel dynamical system, the Dynamical Ising Machine (DIM), which, like the OIM, minimizes the Ising Hamiltonian but offers significantly different dynamical properties. We analyze the characteristic properties of the DIM and compare them with those of the OIM. We also show that the relative performance of each model is dependent on the input graph. Our work illustrates that using multiple dynamical systems with varying properties to solve the same COP enables an effective method that is less sensitive to the input graph, while producing robust solutions.

\end{abstract}

\maketitle

\section{\label{sec:level1}Introduction}

The quest to solve combinatorial optimization problems (COPs) efficiently continues to be a leading-edge challenge in computing owing to the fundamental barriers placed by their computational complexity. For most practical applications, solving problems beyond simple examples necessitates the use of heuristic and meta-heuristic methods. One such method that has attracted recent attention is physics-inspired dynamical systems where the minimization of energy provides a natural analogue to the minimization of the objective function associated with the COP \cite{Wang2021,10011425, PhysRevApplied.17.064064,Vadlamani2020,PhysRevResearch.7.013129,Crnki__2020,Ercsey-Ravasz2011,Molnar2018,shukla2024nonbinarydynamicalisingmachines}. Consequently, as the dynamics evolve towards the ground state, they innately exhibit the ability to compute the solution to the COP.

As a case in point, oscillator Ising machines (OIMs), based on the dynamics of a network of coupled oscillators under second harmonic injection (SHI), exhibit an energy function whose ground state directly corresponds to an optimal configuration of the Ising Hamiltonian,  $H = - \sum_{i,j}^{N} J_{ij} \sigma_i \sigma_j$ \cite{Wang2021}. Here, $J_{ij}$ represents the coupling strength between the spins, $i$ and $j$, $\sigma_i$ and $\sigma_j$ represent the states of the spins $i$ and $j$, respectively, and $N$ is the number of spins in the system. OIMs are being actively explored for solving hard COPs since many COPs can be expressed in terms of minimizing the Ising Hamiltonian \cite{Wang2021,9720612, article,Maher2024, 0ed40711b5fa401a9a92168d17de48f4,moy20221,Goto2021, Albertsson2023,Litvinenko2025, Csaba2020}. Furthermore, OIMs have shown early promise in terms of computational performance (solution quality, time to solution, etc.\cite{Bohm2021}), and consequently motivate further exploration of the physics-inspired approach to solving COPs \cite{erementchouk2024selfcontainedrelaxationbaseddynamicalising}. 

In this work, we propose and evaluate an alternate dynamical system, named Dynamical Ising Machine (DIM), that is also capable of minimizing the Ising Hamiltonian, but exhibits dynamical properties that are starkly different from those of the traditional OIM. With increasing strength of second harmonic injection (SHI), the dynamics of the DIM system exhibit a unique transition in the state corresponding to the lowest energy. Initially, the lowest energy state is at the fixed point $\phi^\star \in \{\frac{\pi}{2}\}$, but as SHI strength increases, the lowest energy state shifts to $\phi^\star \in \{0, \pi\}$. This transition is invariant to the graph topology and manifests as a pitchfork bifurcation in the system's dynamics. (Fig.{~\ref{fig:comparison}}). We also show that the critical SHI strength at the point of bifurcation can provide a good estimate for the ground-state energy. Furthermore, when used in conjunction with the OIM, the differing stability characteristics of the DIM can provide a way to minimize the impact of local minima and help improve the probability of finding the ground state.

Prior to presenting our proposed DIM, we briefly discuss the dynamical properties of the OIM. The OIM dynamics, whose starting point is the Kuramoto model for coupled oscillators, can be expressed as,

\begin{align}
\frac{d\phi_i(t)}{dt} &= -K \cdot \sum_{\substack{j=1 , j \neq i}}^N J_{ij} \cdot \sin\big(\phi_i(t) - \phi_j(t)\big) \notag \\
&\quad 
- K_s \cdot \sin\big(2\phi_i(t)\big), \label{eq:phase}
\end{align}
with a corresponding energy function given by 

\begin{align}
E(\vec{\phi}(t)) &= -K \cdot \sum_{\substack{i,j=1 , j \neq i}}^N J_{ij} \cdot \cos\big(\phi_i(t)  - \phi_j(t)\big)  \notag \\
&\quad 
- K_s \cdot \sum_{i=1}^N \cos\big(2\phi_i(t)\big) .\label{eq:energy}
\end{align}
Here, $K$ is the coupling strength between the oscillators, and $K_s$ is the strength of SHI. As originally shown by Wang \textit{et al.} \cite{Wang2021}, under approporiate level of SHI, the ground state of this dynamical system, represented by an oscillator phase configuration $\phi^\star \in \{0,\pi\}$, maps to a spin configuration $\sigma^\star \in \{+1,-1\}$ that corresponds to the  minimum energy of the Ising Hamiltonian. 

From a dynamical system standpoint, the ground-state phase configurations, $\phi^\star \in \{0,\pi\}$, are fixed points of the OIM. From a computational standpoint, while these are the  `desired' fixed points of the dynamical system, there are many other fixed points which can potentially behave as the local minima and trap the system dynamics resulting in sub-optimal solutions \cite{2023ElL....59E3054B,cheng2024control}. In fact, all spin configurations including those corresponding to higher energies are fixed points of the OIM. Moreover, the relative stability of fixed points corresponding to various spin configurations strongly impacts the computational characteristics of the system, as shown in previous work \cite{Bashar20232}. From the perspective of the system's ability to find the ground state of the Ising Hamiltonian, the desired scenario for stability is that the spin (oscillator phase) configurations corresponding to the ground state are selectively stabilized, whereas all other Ising configurations are unstable. This minimizes the probability of the system dynamics getting trapped into the local minima corresponding to high-energy spin configurations. Moreover, the relative stability of the fixed points is not only a function of the input graph, but also depends on the exact dynamics being used to minimize the objective function. We demonstrate this with the proposed DIM, which, as mentioned earlier, minimizes the Ising Hamiltonian, albeit with different dynamical characteristics.

\section{DIM: An Alternate Dynamical System to Minimize Ising Hamiltonian}

The new dynamical system we propose for minimizing the Ising Hamiltonian is defined as,

\begin{align}
\frac{d\phi_i(t)}{dt} &= -K \cdot \sum_{\substack{j=1 , j \neq i}}^N J_{ij} \cdot \sin\big(\phi_i(t) + \phi_j(t)\big) \notag \\
&\quad 
- K_s \cdot \sin\big(2\phi_i(t)\big).\label{eq:phase2}
\end{align}
It can be observed that the proposed dynamical system uses additive phases (in the first sin(.) term on the RHS) instead of the phase difference $(\phi_i(t) - \phi_j(t))$ used in traditional Kuramoto model-based OIM. Despite the different form, this dynamical system can also minimize the Ising Hamiltonian. To show this, we use an approach similar to \cite{Wang2021}, and consider the  function in Eq.~\eqref{eq:energy2} as the candidate energy function for the dynamics considered in (Eq.~\eqref{eq:phase2})

\begin{align}
E(\vec{\phi}(t)) &= -K \cdot \sum_{\substack{i,j=1 , j \neq i}}^N J_{ij} \cdot \cos\big(\phi_i(t) + \phi_j(t)\big) \notag \\
&\quad 
- K_s \cdot \sum_{i=1}^N \cos\big(2\phi_i(t)\big).\label{eq:energy2}
\end{align}
Equation ~\eqref{eq:energy2} can be shown to be non-increasing i.e., $\frac{dE(\vec{\phi}(t))}{dt} \leq 0$. For this, we first express $\frac{dE(\vec{\phi}(t))}{dt}$ as

\begin{align}
\frac{dE(\vec{\phi}(t))}{dt} = \sum_{k=1}^N \frac{\partial E(\vec{\phi}(t))}{\partial \phi_k} \cdot \frac{d\phi_k(t)}{dt}. \label{eq:dec}
\end{align}
We now evaluate to $\frac {\partial E(\vec{\phi}(t))}{\partial \phi_k} $ in Eq.~\eqref{eq:dec}

\begin{align*}
\frac {\partial E(\vec{\phi}(t))}{\partial \phi_k} &= 
K \cdot \sum_{\substack{l=1 , l \neq k}}^N J_{kl} \cdot \sin\big(\phi_k(t) + \phi_l(t)\big)  \nonumber \\
&\quad 
+K \cdot \sum_{\substack{l=1 , l \neq k}}^N J_{lk} \cdot \sin\big(\phi_l(t) + \phi_k(t)\big) \nonumber \\
&\quad  
+ 2K_s \cdot \sin\big(2 \phi_k(t)\big).
\end{align*}
Noting that $J_{kl} = J_{lk}$, we have

\begin{align*}
\frac {\partial E(\vec{\phi}(t))}{\partial \phi_k} &= 
2 \bigg[K \cdot \sum_{\substack{l=1 , l \neq k}}^N J_{kl} \cdot \sin\big(\phi_k(t) + \phi_l(t)\big)\nonumber \\
&\quad  
+ K_s \cdot \sin\big(2 \phi_k(t)\big)\bigg] .
\end{align*}
From Eq.~\eqref{eq:phase2}

\begin{align*}
\frac {\partial E(\vec{\phi}(t))}{\partial \phi_k} &= 
-2 \cdot \frac{d\phi_k(t)}{dt}.
\end{align*}
Therefore, by substituting $\frac {\partial E(\vec{\phi}(t))}{\partial \phi_k}$ into the Eq.~\eqref{eq:dec}

\begin{align*}
\frac{dE(\vec{\phi}(t))}{dt} &= \sum_{k=1}^N -2 \cdot \frac{d\phi_k(t)}{dt} \cdot \frac{d\phi_k(t)}{dt},
\end{align*}

\begin{align}
\frac{dE(\vec{\phi}(t))}{dt} &= -2\sum_{k=1}^N  \cdot \left(\frac{d\phi_k(t)}{dt}\right)^2 \leq 0. \label{eq:less}
\end{align}
Equation~\eqref{eq:less} indicates that the oscillator dynamics effectively perform gradient descent. Similar to the OIM dynamics, at the phase values of 0 and $\pi$, the energy of the DIM (Eq.~\eqref{eq:less}) is given by equation ~\eqref{eq:ising2}

\begin{align}
E(\vec{\phi}(t)) = -K \sum_{\substack{i,j=1 , j \neq i}}^N J_{ij} \cdot \cos\big(\phi_i(t) + \phi_j(t)\big) 
- N \cdot K_s ,\label{eq:ising2} 
\end{align}
where $- N \cdot K_s $ is a constant. By selecting $K = 1/2$, Eq.~\eqref{eq:ising2} is equivalent to Ising Hamiltonian with a constant offset.

\begin{figure*}[htbp!]

    \centering
    \includegraphics[width=0.98\textwidth]{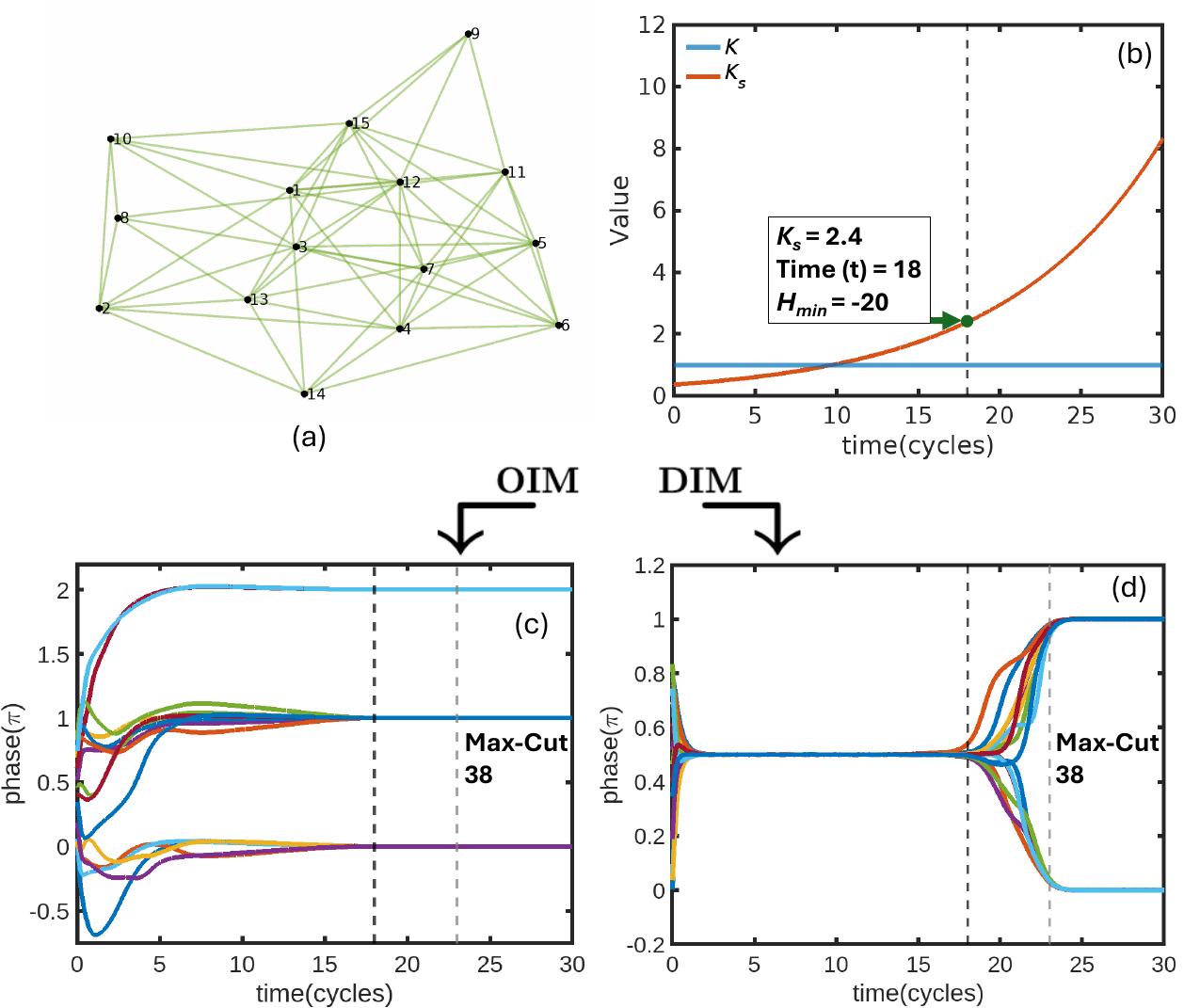}
    \captionsetup{justification=justified} 
    
    \caption{\justifying(a) An Illustrative graph with 15 nodes and 56 edges. The edges were randomly generated. (b) Time evoution of the system parameters, $K (=1)$, and $K_s$ . Resulting phase response of the (c) OIM; and (d) DIM, respectively. In (d), the critical $K_s$ at the point of bifurcation is observed to be $K_{s,E}=2.4$.}   
    
    \label{fig:comparison}
\end{figure*}

We illustrate the DIM characteristics and its computational properties using a randomly generated example graph with 15 nodes and 56 (unweighted) edges (Fig. ~\ref{fig:comparison}). Specifically, here we consider the anti-ferromagnetic Ising Hamiltonian. The solution of the anti-ferromagnetic Ising Hamiltonian directly maps to the  solution of Max-Cut of the graph, an archetypal NP-Hard problem. The Max-Cut problem entails dividing the nodes of the graph into two sets such that the total weight of edges common to both the edges is maximized. The Max-Cut problem can be mapped to the Ising model using the following relationship: $J_{ij}=-W_{ij}$, where $W_{ij}$ is the weight of the edge in the graph and $J_{ij}$ is the corresponding interaction strength between the $i^{th}$ and the $j^{th}$ spin in the Ising model \cite{10.3389/fphy.2014.00005}. The simulations are performed using a stochastic differential equation solver with Gaussian white noise. In the simulated dynamics, we keep the value of $K$ (representing the coupling strength among the oscillators) constant ($K = 1$) while increasing the strength of the SHI, represented by $K_s$, (Fig.\hyperref[fig:comparison]{~\ref*{fig:comparison}(b)}). Additionally, both models used the same randomly generated initial conditions. 

Figures \hyperref[fig:comparison]{~\ref*{fig:comparison}(c),(d)} show the phase evolution observed using the traditional OIM and the DIM, respectively. The phase dynamics observed with the DIM reveal a clear distinction from those of the OIM. Unlike the OIM, the phases in the DIM converge to $\phi \approx \frac{\pi}{2}$ when the strength of the second harmonic injection, $K_s$, is initially small. This can be attributed to $\phi^\star\in \{\frac {\pi}{2}\}$ being the lowest energy configuration at small enough $K_s$, unlike in the case of the OIM, as described further on. In this scenario, the dynamics exhibit reduced dependence on the initial conditions. Subsequently, as $K_s$ increases to a critical value, the phase dynamics then exhibit a pitchfork-like bifurcation. This ultimately drives the phases to either 0 or $\pi$, which can then be mapped to the corresponding spin states. In contrast, no such bifurcation is explicitly observed in the phase dynamics of OIM since the lowest energy phase configuration in the OIM, at low $K_s$, is a function of the graph topology. However, the phases do eventually converge to 0 or $\pi$ once the SHI exceeds a critical threshold \cite{cheng2024control}. For this small illustrative graph, both the models converge to the ground state ($H_{min}=-20)$. The corresponding graph cut, calculated using the relation $H  = -\sum_{\substack{i,j, i \leq j}}^N J_{ij} - 2 \cdot S_{cut}$\cite{10.3389/fphy.2014.00005}, evaluates to 38, which is the Max-Cut for the graph. The optimal value of the cut was verified using BiqMac \cite{RRW10}. \\

\section{Dynamical Properties of DIM}

To understand the observed dynamics of the DIM, we now evaluate its dynamical properties. We specifically focus on understanding the stability of the so-called Type I fixed points, $\phi^\star\in \{\frac {\pi}{2}\}$ and $\phi^\star\in \{0,\pi\}$.  Details of Type I fixed points have been summarized in  
 Appendix~\ref{appendix:proofs}, and detailed in prior work by Chen et al., \cite{cheng2024control}. We first evaluate the local stability of the DIM by analyzing the Lyapunov exponents $(\lambda_1, \lambda_2, \lambda_3, . . ., \lambda_N)$ of various fixed points of interest. This is accomplished by calculating the Eigen values of the Jacobian matrix since it is symmetric \cite{arrowsmith1992dynamical}. For a given phase configuration to be stable, all Lyapunov exponents must be negative indicating that the corresponding Jacobian matrix must be negative-definite.

For the DIM, the Jacobian matrix can be formulated as,

\begin{align}
A_{DIM}(\phi^\star) &= KD_{DIM}(\phi^\star) - 2K_{\rm s}\Delta(\phi^\star)
\label{eq:A_phi}
\end{align}

\begin{widetext}
where,
       \begin{equation*}
          D_{DIM}(\phi^\star )=\begin{bmatrix}
         {\displaystyle -\sum_{j=1,j\ne1}^{N}J_{1j}\cos(\phi_1^\star+\phi_j^\star)} & -J_{12}\cos(\phi_1^\star+\phi_2^\star) & \cdots & -J_{1N}\cos(\phi_1^\star+\phi_N^\star) \\
         -J_{21}\cos(\phi_2^\star+\phi_1^\star) & \displaystyle -\sum_{j=1,j\ne2}^{N}J_{2j}\cos(\phi_2^\star+\phi_j^\star) & \cdots & -J_{2N}\cos(\phi_2^\star+\phi_N^\star)  \\
         \vdots & \vdots & \ddots & \vdots  \\
         -J_{N1}\cos(\phi_N^\star+\phi_1^\star) & -J_{N2}\cos(\phi_N^\star+\phi_2^\star) &  \cdots & \displaystyle-\sum_{j=1,j\ne N}^{N}J_{Nj}\cos(\phi_N^\star+\phi_j^\star)
       \end{bmatrix}\;,\label{Matrix_D} 
        \end{equation*}
       \begin{equation*}
        \Delta(\phi^\star)= \left[\begin{array}{cccc}
         {\cos(2\phi_1^{\star})} & \!\!\!\!\!\!\!\! 0 &\!\!\!\!\!\!\!\!\cdots & \!\!\!\!\!\!\!\!0 \\
         \!\!\!\!\!\!\!\!0 & \!\!\!\!\!\!\!\!\cos(2\phi_2^\star)&\cdots & \!\!\!\!\!\!\!\!0  \\
         \!\!\!\!\!\!\!\!\vdots& \!\!\!\!\!\!\!\!\vdots& \!\!\!\!\!\!\!\!\ddots & \!\!\!\!\!\!\!\! \vdots\\
        \!\!\!\!\!\!\!\! 0 & \!\!\!\!\!\!\!\!0 &\!\!\!\!\!\!\!\!\cdots & \!\!\!\!\!\!\!\!\cos(2\phi_N^\star)
       \end{array}\!\!\right] 
       \end{equation*}
\end{widetext}

Using Equation~\eqref{eq:A_phi}, we now analyze the local stability of the DIM at Type I fixed points, and compare their properties to those of the OIM.

\subsection{Characteritics of \boldmath$\phi^\star \in \left\{\frac{\pi}{2}\right\}$  fixed points}

We first investigate the properties of the fixed point $\phi^\star \in \left\{\frac{\pi}{2}\right\}$. In the absence of second harmonic injection {i.e., $K_{s}=0$), $\phi^\star \in \left\{\frac{\pi}{2}\right\}$ is the lowest energy ($E_{DIM}= K \cdot \sum_{\substack{i,j=1 , j \neq i}}^N J_{ij} $; Eq.~\eqref{eq:ising2}) configuration of the DIM since $ \cos\big(\phi^\star + \phi_j^\star\big) = -1$ for all the edges in the network. This is in contrast to the OIM, where the fixed point $\phi^\star \in \left\{\frac{\pi}{2}\right\}$ corresponds to the largest possible system energy, given by $E_{OIM}= -K \cdot \sum_{\substack{i,j=1 , j \neq i}}^N J_{ij} $, when $K_s=0$. This minimizes the probability of realizing this configuration in practical systems. We emphasize here that these observations on the relative energy of the fixed point $\phi^\star \in \left\{\frac{\pi}{2}\right\}$ is based on the anti-ferromagnetic Ising Hamiltonian, relevant to the solving the Max-Cut COP. For the ferromagnetic case $(W_{ij}=J_{ij})$, the scenario is reversed wherein the  $\phi^\star \in \left\{\frac{\pi}{2}\right\}$ lies at the highest energy in the DIM whereas it is lies at the lowest energy in the OIM.\\

Next, we analyze the local stability of this phase configuration. At the fixed point $\phi^\star \in \left\{\frac{\pi}{2}\right\}$ $(\equiv \phi^{\star\mu})$, the Jacobian matrix, defined in Eq.~\eqref{eq:A_phi} reduces to, 

\begin{align}
A_{DIM}(\phi^{\star\mu}) = KD_{DIM}(\phi^{\star\mu}) + 2K_{\rm s}I_N,
\label{eq:equinewpi_2}
\end{align}
where, $I_N$ refers to an identity matrix of size $N \times N$. Furthermore, $D_{DIM}(\phi^{\star\mu})$ is negative semi-definite (see Appendix~\ref{appendix:negative-semi}) indicating that the system is stable at $\phi^{\star\mu}$, although asymptotic stability can only be guaranteed in graph instances where $D_{DIM}(\phi^{\star\mu})$ is negative definite. Subsequently, the SHI signal, described by the second term in Eq. ~\eqref{eq:equinewpi_2}, then serves to progressively destabilize this phase configuration in addition to increasing the system energy at this phase configuration. 

With regards to the observed bifurcation in the DIM, the critical SHI signal strength ($K^{\{\frac{\pi}{2}\}}_{s,ls}$) at which $\phi^\star \in \left\{\frac{\pi}{2}\right\}$ fixed point becomes unstable can be determined as follows. Let $\lambda_L(D_{DIM}(\phi^{\star\mu})$ be the largest Lyapunov exponent of  $D_{DIM}(\phi^{\star\mu})$. Then, from equation~\eqref{eq:equinewpi_2}, all eigen values of $A_{DIM}(\phi^{\star\mu})$ are negative if $K_{\rm s}<\frac{-K\lambda_L(D_{DIM}(\phi^{\star\mu}))}{2}$. Therefore, the critical SHI signal strength ($K^{\{\frac{\pi}{2}\}}_{s,ls}$) required to destabilize this fixed point can be formulated as,

\begin{subequations}
\begin{align}
K^{\{\frac{\pi}{2}\}}_{s,ls}=\frac{-K\lambda_L(D_{DIM}(\phi^{\star\mu}))}{2}.
\label{eq:Ks_critical_half_pi}
\end{align}
Furthermore, Eq.~\eqref{eq:Ks_critical_half_pi} can expressed as, 

\begin{align}
K^{\{\frac{\pi}{2}\}}_{s,ls}=\frac{-K\lambda_L(-D - W)}{2},
\label{eq:Ks_critical_half_pi2}
\end{align}
\end{subequations}
since it can be shown that at $\phi^\star \in \left\{\frac{\pi}{2}\right\}$, $D_{DIM}(\phi^{\star\mu}) = -D-W$, where, $D$ is the degree matrix and $W$ is the weight matrix.

In the following section, we will show that the SHI signal has the opposite effect on the energy and the local stability of $\phi^\star \in \{0,\pi\}$ fixed points. Consequently, in the presence of perturbations, as $K_{\rm s}$ increases, the system will eventually exhibit a bifurcation when other configurations become energetically more favorable than the $\phi^\star \in \left\{\frac{\pi}{2}\right\}$. 

\subsection{Characteristics of \boldmath$\phi^\star \in \{0,\pi\}$ fixed points}
We now analyze the properties of the fixed points $\phi^\star \in \{0,\pi\}$ of the DIM. The Jacobian matrix for $\phi^\star \in \{0,\pi\}$ $(\equiv \phi^{\star\nu})$ fixed points can be derived from  Eq.~\eqref{eq:A_phi} as,

\begin{align}
A_{DIM}(\phi^{\star\nu}) = KD_{DIM}(\phi^{\star\nu}) - 2K_{\rm s}I_N
\label{eq:equinewpi}
\end{align}
As alluded to in the previous section, increasing the SHI strength ($K_{s}$ in equation Eq.~\eqref{eq:equinewpi}) increases the stability and reduces the energy (see Eq.~\eqref{eq:energy2}) of the fixed points $\phi^\star \in \{0,\pi\}$. 

Next, we define some critical thresholds with regards to the bifurcation process. Similar to case of $\phi^{\star\mu}$, a critical threshold can be defined for the SHI strength at at which $\phi^\star \in \{0,\pi\}$ fixed point becomes stable. Using Eq. ~\eqref{eq:equinewpi}, the critical SHI signal ($K^{\{0,\pi\}}_{s,ls}$) at which $\phi^\star \in \{0,\pi\}$ fixed point becomes stable can defined as, 

\begin{align}
K^{\{0,\pi\}}_{s,ls}=\frac{K\lambda_L(D_{DIM}(\phi^{\star\nu}))}{2},
\label{eq:Ks_critical_zero_pi}
\end{align}
where, $\lambda_L(D_{DIM}(\phi^{\star\nu}))$ is the largest Lyapunov exponent of  $D_{DIM}(\phi^{\star\nu})$. 

Therefore, as mentioned earlier, with increasing $K_s$, the $\phi^{\star\mu}$ become increasingly unstable, and stable respectively, manifesting as a bifurcation. We also consider the critical SHI signal strength $K_{s,E}$ at which the $\phi^\star \in \{0,\pi\}$ fixed points, at the lowest energy, become energetically more favorable compared to the $\phi^\star \in \left\{\frac{\pi}{2}\right\}$ fixed points using Eq.~\eqref{eq:commonlabel}. 
\begin{subequations}
\label{eq:commonlabel}
\begin{align}
-K \cdot \sum_{\substack{i,j=1 , j \neq i}}^N J_{ij} \cdot \cos\big(\phi_i^{\star\nu} + \phi_j^{\star\nu}\big)  
 - K_{s,E} \cdot \sum_{i=1}^N \cos\big(2\phi_i^{\star\nu}\big) & = \nonumber \\
-K \cdot \sum_{\substack{i,j=1 , j \neq i}}^N J_{ij} \cdot \cos\big(\phi_i^{\star\mu} + \phi_j^{\star\mu}\big)
- K_{s,E} \cdot \sum_{i=1}^N \cos\big(2\phi_i^{\star\mu}\big). \label{eq:energycriticala}
\end{align}
At the respective fixed points, $\phi^{\star\nu}  \in \{0,\pi\}$ and $\phi^{\star\mu} \in \left\{\frac{\pi}{2}\right\}$, the equation Eq.~\eqref{eq:energycriticala} reduces to, 
\begin{align*}
-K \cdot \sum_{\substack{i,j=1 , j \neq i}}^N J_{ij} \cdot \cos\big(\phi_i^{\star\nu} + \phi_j^{\star\nu}\big)  
 - K_{s,E} \cdot N & = \nonumber \\
-K \cdot 
\sum_{\substack{i,j=1 , j \neq i}}^N J_{ij} \cdot(-1)
+ K_{s,E} \cdot N .
\end{align*}
Which can then be used to express $K_{s,E}$ as shown below in Eq.~\eqref{eq:energycriticalb}, 
\begin{align} 
K_{s,E}  =\frac{ 2 K \cdot \big(-\sum_{\substack{i,j, i \leq j}}^N J_{ij} \cdot \cos\big(\phi_i^{\star\nu} + \phi_j^{\star\nu}\big)
- \sum_{\substack{i,j, i \leq j}}^N J_{ij}\big)}{ 2 \cdot N}.\label{eq:energycriticalb}
\end{align}
The term $-\sum_{\substack{i,j, i \leq j}}^N J_{ij} \cdot \cos\big(\phi_i^{\star\nu} + \phi_j^{\star\nu}\big)$ in Eq.~\eqref{eq:energycriticalb}, corresponds to the globally optimal solution of the Ising Hamiltonian which we refer to as $H_{min}$, and $-\sum_{\substack{i,j, i \leq j}}^N J_{ij}$ corresponds to the total number of edges ($\xi$) in the graph. Thus, Eq.~\eqref{eq:energycriticalb} reduces to, 

\begin{align} 
K_{s,E}  =\frac{ K \cdot \big( H_{min} 
+ \xi \big)}{N}.\label{eq:energycriticalc}
\end{align}
Alternatively, $H_{min}$ can be expressed in terms of $K_{s,E}$ as shown in Eq.~\eqref{eq:estimation}, 

\begin{align} 
H_{min}  = \frac{K_{s,E}\cdot N}{K} - \xi. \label{eq:estimation} 
\end{align}
\end{subequations}

It can be observed that among the three critical thresholds,  $K^{\{\frac{\pi}{2}\}}_{s,ls}$, $K^{\{0,\pi\}}_{s,ls}$,  and $K_{s,E}$ Eq.~\eqref{eq:energycriticalc}, $K^{\{0,\pi\}}_{s,ls}$ is relevant to the onset of the DIM dynamics behaving as an Ising machine. However, computing $K^{\{0,\pi\}}_{s,ls}$ is challenging it requires prior knowledge of the ground state configuration. In contrast, we observe empirically that approximating $K_{s,E}$ to the point of bifurcation obtained from the dynamics can provide high quality estimates of the solution.

We first illustrate this using the graph considered in Fig. \hyperref[fig:comparison]{~\ref*{fig:comparison}(a)}. From Fig.\hyperref[fig:comparison]{~\ref*{fig:comparison}(d)}, the critical $K_s$ can be estimated to be $K_{s,E} \approx 2.4$ (at $t \approx 18$), which can subsequently be used to calculate $H_{min} (=-20)$ and the corresponding cut (=38), which is the optimal solution for the graph. We use a threshold ($\Delta=0.006$) to determine the bifurcation point as discussed in Appendix~\ref{appendix:kscritical}.

\begin{table}[h!]
\centering
\caption{\justifying Estimated cut for the G1 to G5 instances from the G-set benchmark \cite{Gset}. The cut is calculated using the Ising energy (rounded off to the nearest integer), estimated using the value of $K_{s,E}$ obtained from the bifurcation in the DIM dynamics. The estimated values are also compared with best known cuts for the graphs.}
\begin{tabular}{|c|c|c|c|c|c|}
\hline
\multicolumn{3}{|c|}{\textbf{Graph Information}} & \textbf{Estimated} & \textbf{Target} & \textbf{Ratio} \\ \hline
\textbf{\#} & \textbf{Nodes} & \textbf{Edges} 
 & \textbf{Max-Cut} & \textbf{Max-Cut} & \textbf{\%} \\ \hline

G1 & 800 & 19,176 &  11,623 & 11,624 & 99.99 \\ \hline
G2 & 800 & 19,176 &  11,620 & 11,620 & 100 \\ \hline
G3 & 800 & 19,176 &  11,620 & 11,622 & 99.98 \\ \hline
G4 & 800 & 19,176 &  11,629 & 11,646 & 99.85 \\ \hline
G5 & 800 & 19,176 &  11,604 & 11,631 & 99.77 \\ \hline

\end{tabular}
\label{tab:gset}
\end{table}

We also test this method approach on few instances (G1-G5) from the G-set benchmark. Table~\ref{tab:gset}. shows the value of the cut computed using $K_{s,E}$ obtained from the bifurcation dynamics of the DIM and compared  to the best known cuts for the graphs. The same threshold ($\Delta$) was used across all instances. It can be observed that in all cases, the DIM computes a cut that is at least 99.77\% of the best-known value. We do note here that the quality of the resulting solution calculated using this method will be sensitive to the value of $K_{s,E}$ derived from the dynamics. Nevertheless, this method can help provide a range where the minimum solution is expected to lie.

\subsection{Local Stability Properties of DIM and OIM}

Next, we establish that the relative local stability of the $\phi^\star \in \{0,\pi\}$ phase configurations, as a function of energy, can be different for the OIM and the DIM. This implies that the choice of the computational model can alter the relative stability of various phase configurations, which in turn, can be a useful knob to improve the overall effectiveness of physics-inspired methods. As illustrated in prior work \cite{Bashar20232}, the relative stability of the ground state configurations with respect to solutions that are energetically close but sub-optimal directly affects the probability of finding the ground state solution. 

To evaluate the local stability properties of the DIM and OIM, we calculate the Lyapunov exponents of their respective Jacobian matrices. The Jacobian matrix for the DIM ($A_{DIM}$ Eq.~\eqref{eq:equinewpi}) can be expressed in terms of $A_{OIM}$, formulated for the OIM in our prior work \cite{cheng2024control}, as 
\begin{align}
A_{DIM}(\phi^\star) = A_{OIM}(\phi^\star) - 2 \aleph(\phi^\star )
\label{eq:equi3}
\end{align}
where, \begin{widetext}
\begin{equation*}
\aleph(\phi^\star )=\begin{bmatrix}
{0} & J_{12}\cos(\phi_1^\star-\phi_2^\star) & \cdots & J_{1N}\cos(\phi_1^\star-\phi_N^\star) \\
J_{21}\cos(\phi_2^\star-\phi_1^\star) & 0 & \cdots & J_{2N}\cos(\phi_2^\star-\phi_N^\star)  \\
\vdots & \vdots & \ddots & \vdots  \\
J_{N1}\cos(\phi_N^\star-\phi_1^\star) & J_{N2}\cos(\phi_N^\star-\phi_2^\star) &  \cdots & 0 \end{bmatrix}\;\label{zeta} 
\end{equation*}
\end{widetext}

Equation~\eqref{eq:equi3} reveals that the DIM and the OIM have distinct Jacobian matrices. Moreover, since $\aleph$ depends on the fixed point $\phi^\star \in \{0,\pi\}$ where the stability is evaluated and the input graph (through the adjacency matrix, $J$), it also implies that the relative stability of various fixed points is dependent on the choice of the dynamical system. Specifically, local stability of fixed points at different energies may vary depending on the graph and the choice of the dynamical system (here, DIM or OIM) making some graphs more suitable to a given dynamical system over the other.

\section{Impact of Dynamical System on Computational Performance}

\begin{figure*}[htbp!]

    \centering
    \begin{subfigure}[b]{0.30\textwidth}
        \centering
        \captionsetup{labelformat=empty}
        \includegraphics[width=\textwidth]{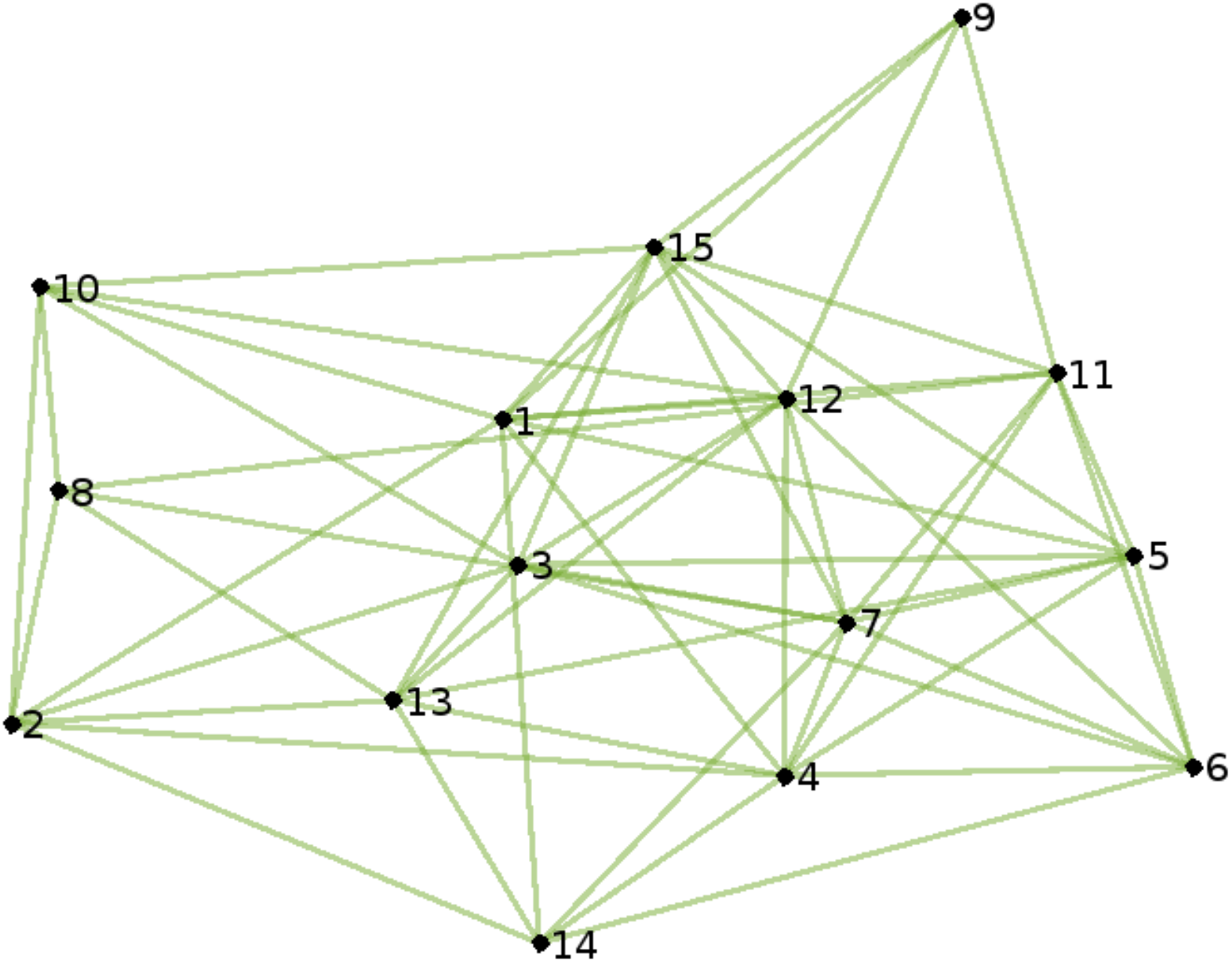}
        \caption{(a) Graph 1}
        \label{fig:suba}
    \end{subfigure}
    \hspace{0.12\textwidth} 
    \begin{subfigure}[b]{0.30\textwidth}
        \centering
        \captionsetup{labelformat=empty}
        \includegraphics[width=\textwidth]{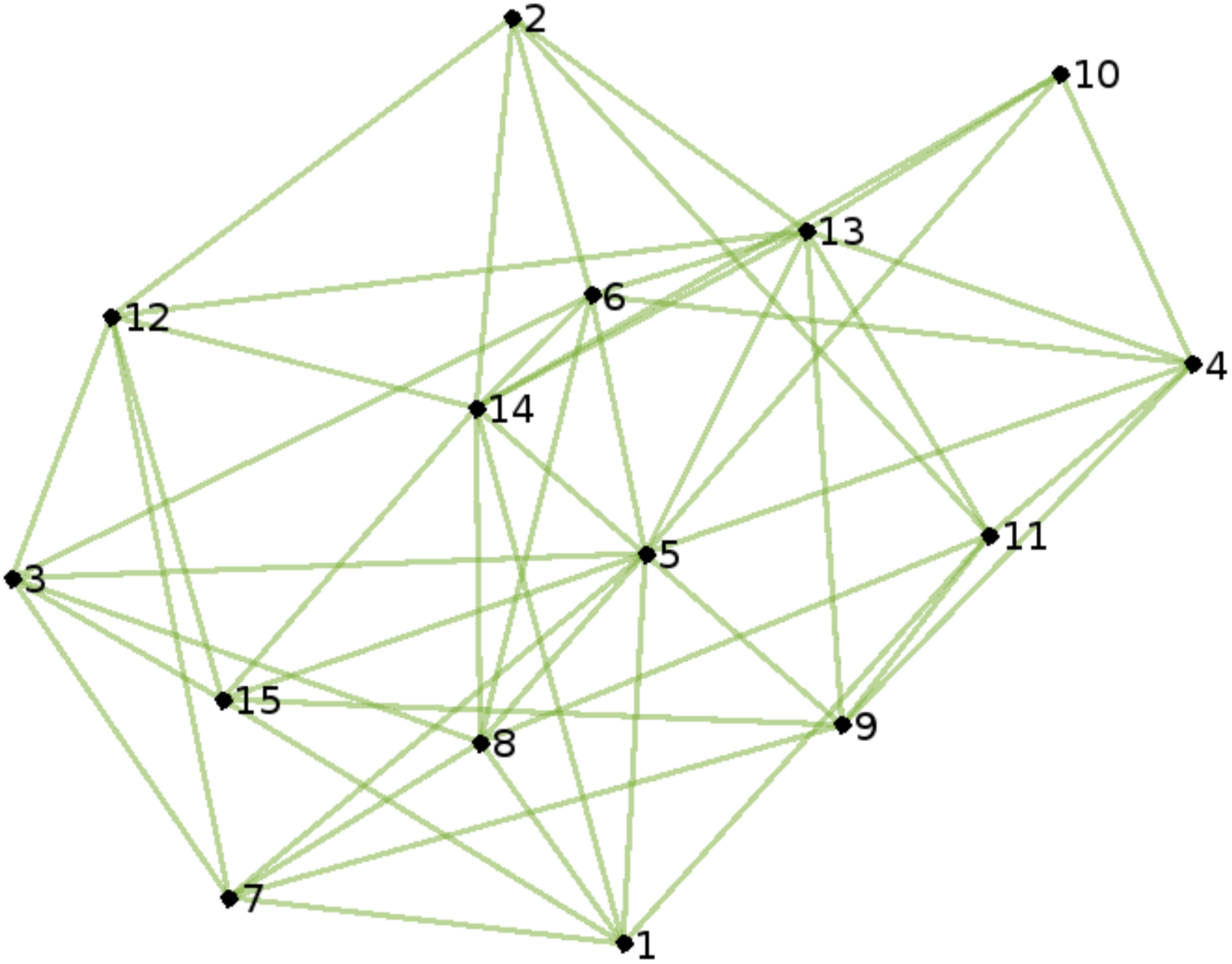}
        \caption{(b) Graph 2}
        \label{fig:subb}
    \end{subfigure}

    \begin{subfigure}[b]{0.48\textwidth}
        \centering
        \captionsetup{labelformat=empty}
        \captionsetup{justification=centering}
        \caption{
        \parbox{\textwidth}{ 
            \raisebox{0.5ex}{\textsmaller[1]{}} \\ 
            \raisebox{0ex}{\small \textbf{OIM}} \hspace{20pt} \raisebox{0ex}{\small \textbf{DIM}} \\
            \scalebox{5}{\rotatebox[origin=c]{90}{$\Lsh$}}  
            \rule{20pt}{0pt} 
            \hspace{3pt} 
            \rule{20pt}{0pt} 
            \scalebox{5}{\rotatebox[origin=c]{270}{$\Rsh$}}}}
    \end{subfigure}
    \begin{subfigure}[b]{0.48\textwidth}
        \centering
        \captionsetup{labelformat=empty}
        \captionsetup{justification=centering}
        \caption{
        \parbox{\textwidth}{ 
            \raisebox{0.5ex}{\textsmaller[1]{}} \\ 
            \raisebox{0ex}{\small \textbf{OIM}} \hspace{20pt} \raisebox{0ex}{\small \textbf{DIM}} \\ 
            \scalebox{5}{\rotatebox[origin=c]{90}{$\Lsh$}}  
            \rule{20pt}{0pt} 
            \hspace{3pt} 
            \rule{20pt}{0pt} 
            \scalebox{5}{\rotatebox[origin=c]{270}{$\Rsh$}}}}       
   \end{subfigure}

        \begin{subfigure}[b]{\textwidth}
            \centering
            \captionsetup{labelformat=empty, skip=0pt}
            \includegraphics[width=\textwidth]{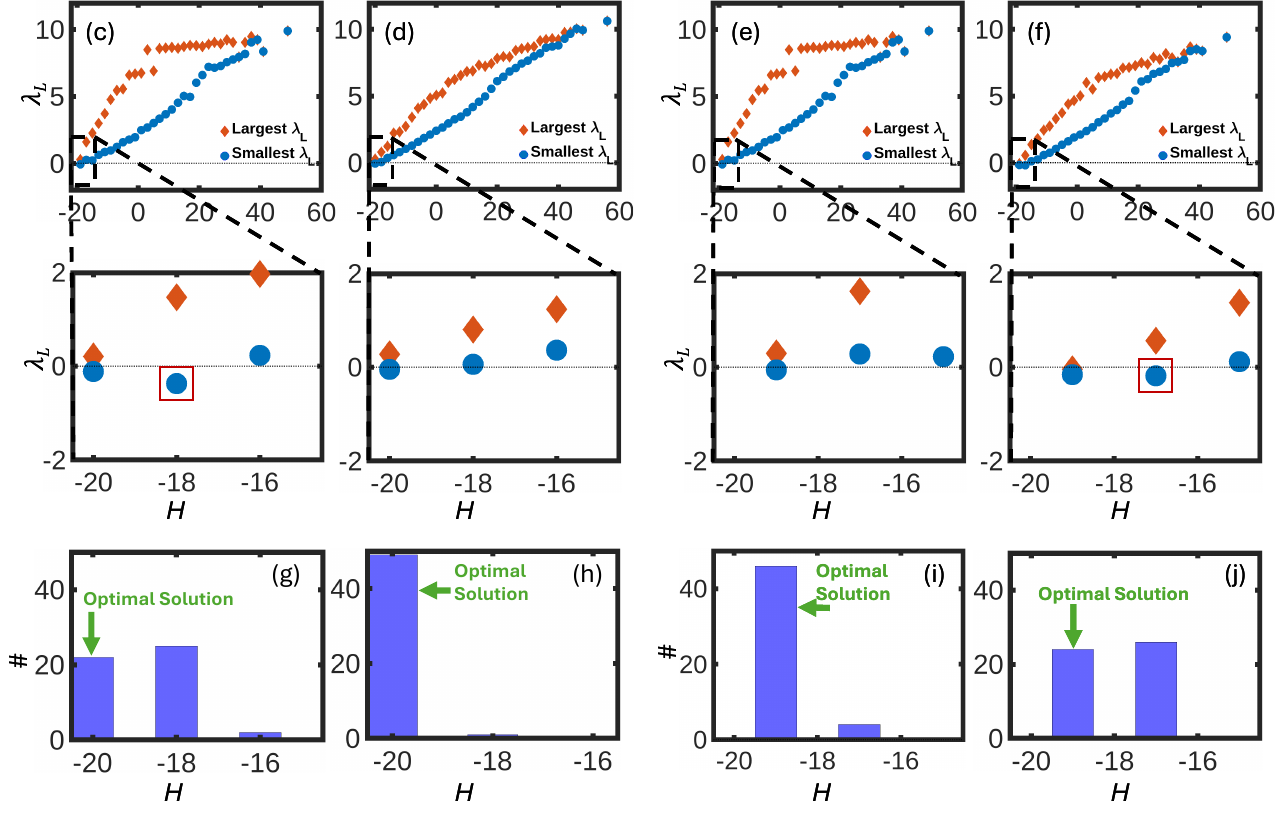}
            \label{fig:graph1_o}
        \end{subfigure} 
        \vspace{-25pt}
    \captionsetup{justification=justified} 
    \caption{\justifying (a)(b) Illustrative graphs (graph \#1 and graphs\#2) with 15 nodes considered in the example, respectively. (c-f) Minimum (blue circles) and maximum (orange diamonds) value of $\lambda_L$ for the various phase configurations at each Ising energy, calculated using the OIM and the DIM dynamics, for each of the two graphs, respectively. Insets magnifying the relative distribution of $\lambda_L$ as a function of energy close to the ground state are also shown. (g-j) Histogram of the Ising energy ($H$) computed using the OIM and the DIM dynamics for graph \#1 and \#2, respectively, for 50 independent trials.}
    \label{fig:rand15}
\end{figure*}

To illustrate how the choice of the  dynamical system (here, OIM and DIM) can impact the computational performance, we consider two illustrative graphs (graph 1 and graph 2) with 15 nodes, as shown in Fig. ~\ref{fig:rand15}. The connectivity is randomly generated with graph 1 consisting of 56 edges, and graph 2 having 49 edges. For every fixed point $\phi^\star \in \{0,\pi\}$, we calculate the Lyapunov exponents of the Jacobian matrix for the OIM and the DIM dynamics, for each graph (1 and 2). Subsequently, we analyze the largest Lyapunov exponent, $\lambda_{L}$, for all the possible phase configurations as a function of their energy. Specifically, since multiple phase configurations can have the same energy, we focus on the largest and the smallest $\lambda_{L}$ (i.e., range of $\lambda_{L}$) at a given energy. Figs.\hyperref[fig:rand15]{~\ref*{fig:rand15}(c-f)},  show the maximum ($\lambda_{L,max}$; orange diamonds) and the minimum ($\lambda_{L,min}$; blue circles) values of $\lambda_L$ for each combination of the dynamical system (DIM, OIM) and the graph (graph 1 and 2). For a phase configuration to be stable, $\lambda_L < 0$, thus ensuring that all the Lyapunov exponents are negative.

For the simulations performed in Fig.~\ref{fig:rand15}, the system parameters, $K$ and $K_s$, are specifically chosen such that at least some of the ground-state phase configurations are stabilized while ensuring that the maximum number of fixed points $\phi^\star \in \{0,\pi\}$ lying at higher energies remain unstable. This ensures that: (a) At least some of the phase configurations, corresponding to optimal solutions of the Ising Hamiltonian, are attractive fixed points of the dynamical systems; (b) The chance of the system dynamics getting trapped in the high energy fixed points (corresponding to sub-optimal solutions) is minimized. However, as evident from Fig.~\ref{fig:rand15}, selectively stabilizing the lowest energy fixed points may not always be possible. For example, if $\lambda_{L,min}$ for the ground state is larger than $\lambda_{L,min}$ for phase configurations with higher energies, it implies that the ground state cannot be selectively stabilized. In other words, at least some of the sub-optimal configurations will be stabilized before the ground state can become stable. Consequently, these higher energy configurations can then behave as local minima, capable of trapping the dynamics (if and when the dynamics fall within the domain of attraction of that fixed point) and resulting in sub-optimal solutions.

This scenario is realized in Fig.\hyperref[fig:rand15]{~\ref*{fig:rand15}(c)} when considering the OIM model to compute the minimum Ising energy for graph 1, and in Fig.\hyperref[fig:rand15]{~\ref*{fig:rand15}(f)}  considering the DIM model to compute the minimum Ising energy for graph 2. However, in the complementary cases i.e., using DIM to minimize the Ising energy of graph 1, and OIM to minimize the Ising energy of graph 2, it can be observed that the ground states can be selectively stabilized (Fig.\hyperref[fig:rand15]{~\ref*{fig:rand15}(d)} and Fig.\hyperref[fig:rand15]{~\ref*{fig:rand15}(e)}). These examples help showcase the fact that the local stability properties are a strong function of the graph, $J$, and the dynamical system considered. 

Furthermore, as alluded to earlier, these differences in relative stability also get reflected in the computational performance of the two dynamical systems when solving COPs. We evaluate this impact in the case of the two graphs considered above by performing 50 independent trials with each dynamical system (OIM, DIM) for each of the two graphs. The initial conditions for a trial are randomly generated but are common to the OIM and the DIM. Subsequently, the histogram of the Ising energy ($H$) obtained for each combination of graph and dynamical system (DIM, OIM) is shown in Figs.\hyperref[fig:rand15]{~\ref*{fig:rand15}(g-j)}. It can observed that in cases where the ground state phase configurations can be selectively stabilized (Fig.\hyperref[fig:rand15]{~\ref*{fig:rand15}(d)} and Fig.\hyperref[fig:rand15]{~\ref*{fig:rand15}(e)}), the probability of the corresponding dynamical system finding the ground state is larger than in scenarios where the ground state cannot be stabilized. The degraded solution quality in the latter case can likely be attributed to the system dynamics getting trapped in the local minima (attractive fixed points) lying at higher energies.

However, we note that while employing the appropriate dynamical system to solve a given graph can improve the chance of finding the ground state, identifying the most suitable model \textit{a priori}, is challenging. Consequently, for solving practical COPs effectively using dynamical systems, we propose employing a `diversification' strategy that utilizes multiple parallel trials that employ different dynamical systems to minimize the same objective function. The concept of using multiple trials with differing initial conditions has been commonly used in other algorithms such as simulated annealing \cite{G.10}. However, in our proposed method, we aim to go one step beyond by allowing the system to minimize a given objective function using a variety of dynamics (here, DIM and OIM) in order to improve the chances of finding to the ground state. 

\begin{table}[h!]
\centering
\caption{\justifying Comparison of the Max-Cut computed using the DIM and OIM for various instances of 50 node graphs with randomly generated connectivity. The frequency with which the ground state was obtained is also reported. For each graph, 50 independent trials were performed using randomly generated initial conditions that are common to the OIM and the DIM. The optimal Max-Cut was verified using the BiqMac solver \cite{RRW10}.}
\begin{tabular}{|c|c|c|c|c|cc|cc|}
\hline
\multicolumn{4}{|c|}{\textbf{Graph Information}} & \textbf{Target} & \multicolumn{2}{c|}{\textbf{OIM}} & \multicolumn{2}{c|}{\textbf{DIM}} \\ \hline
\textbf{\#} & \textbf{Nodes} & \textbf{Edges} & \textbf{Weight} & \textbf{Max-Cut} & \textbf{Cut} & \textbf{\#} & \textbf{Cut} & \textbf{\#} \\ \hline

1 & 50 & 126 & +1 & 100 & 100 &\textbf{40} & 100 & 29 \\ \hline
2 & 50 & 130 & +1 & 104 & 104 & 46 & 104 & \textbf{50} \\ \hline
3 & 50 & 236 & +1 & 167 & 167 & \textbf{35} & 167 & 16 \\ \hline
4 & 50 & 261 & +1 & 181 & 181 & 40 & 181 & \textbf{48} \\ \hline
5 & 50 & 491 & +1 & 299 & 299 & \textbf{32} & 299 & 19  \\ \hline
6 & 50 & 495 & +1 & 315 & 315 & 33 & 315 & \textbf{50} \\ \hline
\end{tabular}
\label{tab:split_four}
\end{table}

To illustrate this method, we test our approach on several randomly generated graph instances, as shown in Table~\ref{tab:split_four}. Each graph is evaluated using 50 independent trials for both the DIM and the OIM. The initial conditions for a trial are generated randomly but common to the OIM and the DIM. The noise amplitude is kept the same in both the systems. The maximum value of the cut computed by each method along with the number of trials where the maximum cut was obtained is shown in Table~\ref{tab:split_four}. It can be observed that while both the methods are able to compute the ground state (confirmed using the Biq-Mac solver \cite{RRW10}), the probability of finding it varies for each system and depends on the graph. As discussed earlier, these results suggest that the relative performance of each model is highly dependent on the input graph structure. Consequently, a unified framework integrating multiple dynamical systems can help produce consistently high-quality solutions across a diverse range of graphs while minimizing the sensitivity of a specific dynamical system to a specific graph. This also encourages the exploration of alternate dynamical systems beyond the DIM that minimize the Ising Hamiltonian but offer different dynamical properties.

\section{Conclusion}

This paper introduces a new dynamical system, DIM, which, while similar to the OIM in its ability to minimize the Ising Hamiltonian, exhibits drastically different dynamical properties. Our work  not only reveals a novel facet of physics-inspired methods where different dynamical systems accomplishing the same computational objective can offer different dynamical properties, but also demonstrates that this diversification in the dynamics can provide a useful toolkit to improve the computational performance. Our findings not only encourage the exploration of new dynamical systems beyond the DIM, but also motivate research into the physical implementation of such novel dynamics.

\section*{Acknowledgments}
The authors would like to thank Professor Zongli Lin (UVA), Yi Cheng (UVA) for valuable feedback on the manuscript, and Nikhat Khan (UVA) for support with the codes. This material was based upon work supported by the National Science Foundation (NSF) under Grant No. 2328961 and was supported in part by funds from federal agency and industry partners as specified in the Future of Semiconductors (FuSe) program.\\

\textbf{Author Contributions:} \textbf {E.M.H.E.B Ekanayake:} Conceptualization (equal); Formal analysis (equal); Software (equal); Writing – original draft (equal); Writing – review \& editing (equal). \textbf {Nikhil Shukla:} Conceptualization (equal); Funding acquisition (lead); Supervision (lead); Validation (equal); Writing – review \& editing (equal).

\renewcommand{\appendixname}{\MakeUppercase{Appendix}}
\appendix
\section{\MakeUppercase{Definition of Type I Fixed Points}}

\textbf{Definition:} A Type I fixed point $\phi^\star =col \{\phi_1^\star, \phi_2^\star, . . ., \phi_N^\star\}$ satisfies $\phi_i^\star\in \{\frac {k\pi}{2}: k \in \mathbb{Z}\}$ for all \textit{i}. Most importantly, Type I fixed points remain invariant to the system parameters and topology.

\textbf{Theorem:} Let $\phi^\star=col \{\phi_1^\star, \phi_2^\star, . . ., \phi_N^\star\} \in \mathbb{R}^N$ be a Type I fixed point of Eq.~\eqref{eq:phase2}. Then either $\phi_i^\star\in \{\frac {k\pi}{2}: \frac {k}{2} \in \mathbb{Z}\}$ for all \textit{i} or $\phi_i^\star\in \{\frac {k\pi}{2}: \frac {k+1}{2} \in \mathbb{Z}\}$ for all \textit{i}.

\textbf{Proof:} The fixed point $\phi^\star$ satisfies the phase dynamics of Eq.~\eqref{eq:phase2}, that is,

\begin{align}
 K \cdot \sum_{\substack{j=1 , j \neq i}}^N J_{ij} \cdot \sin\big(\phi_i^\star + \phi_j^\star\big) = 
- K_s \cdot \sin\big(2\phi_i^\star\big). \label{eq:proof}
\end{align}
Since $\phi^\star$ remains unchanged despite variations in the system parameters, Eq.~\eqref{eq:proof} holds for all values of $K$ and $K_s$, which implies that

\begin{equation}
\begin{cases}
    \sin\big(2\phi_i^\star\big) =  0, \quad i = 1,2,...,N,  
 \\[10pt]
    \sum_{\substack{j=1 , j \neq i}}^N J_{ij} \cdot \sin\big(\phi_i^\star + \phi_j^\star\big)
    = 0, \quad i = 1,2,...,N.
\end{cases}
\label{eq:condition2}
\end{equation}
In addition, since Eq.~\eqref{eq:condition2} holds for all possible values of $J_{ij}$, we have

\begin{equation*}
\begin{cases}
    2\phi_i^\star =  k\pi,\quad i = 1,2,...,N , 
 \\[10pt]
    \phi_i^\star + \phi_j^\star = k\pi,\quad i = 1,2,...,N,
\end{cases}
\label{eq:condition3}
\end{equation*}
where $k\in  \mathbb{Z}$. It then follows that either 
$\phi_i^\star\in \{\frac {k\pi}{2}: \frac {k+1}{2} \in \mathbb{Z}\}$ for all $i$ or $\phi_i^\star\in \{\frac {k\pi}{2}: \frac {k}{2} \in \mathbb{Z}\}$ for all $i$. 

Consequently, this implies that a Type I fixed point $\phi^\star =col \{\phi_1^\star, \phi_2^\star, . . ., \phi_N^\star\}$ satisfies $\phi_i^\star\in \{\frac {k\pi}{2}: k \in \mathbb{Z}\}$ for all \textit{i}.

\label{appendix:proofs}

\section{\MakeUppercase{Negative Semi-Definiteness of \boldmath$D_{DIM}(\phi^\star)$ at $\phi^\star \in \left\{\frac{\pi}{2}\right\}$}}

\textbf{Proof:} For $D_{DIM}(\phi^\star)$ to be negative semi-definite, the following relationship should hold for any vector \( \mathbf{x} \in \mathbb{R}^n \)\cite{Horn2012},

\begin{equation}
\mathbf{x}^T D_{\text{DIM}}(\phi^\star) \mathbf{x} \leq 0 \label{eq:quad} .
\end{equation}
At the fixed point $\phi^\star \in \left\{\frac{\pi}{2}\right\}$, $D_{DIM}(\phi^\star)$ exhibits the following relationship,

\begin{equation*}
D_{DIM}(\phi^\star) = - D - W,
\end{equation*}
where, $D$ is the diagonal matrix where each entry represents the degree of a node 
\[
  D_{ii} = \sum_{j} W_{ij},
\]
and $W$ is the weight matrix where off-diagonal elements represent the edge weights of the graph
  \[
  W_{ij} =
  \begin{cases}
    1, & \text{if there is an edge between nodes } i \text{ and } j, \\
    0, & \text{otherwise}.
  \end{cases}
  \]
We note that the weight matrix \( W \) is a symmetric matrix.  \\

Thus, for any vector \( \mathbf{x} \), Eq.~\eqref{eq:quad} can be expressed as follows,

\begin{subequations}
\begin{equation}
\mathbf{x}^T D_{DIM}(\phi^\star) \mathbf{x} = \mathbf{x}^T (-D - W) \mathbf{x}. \label{eq:quadnew}
\end{equation}
Expanding Eq.~\eqref{eq:quadnew}

\begin{equation}
\mathbf{x}^T D_{\text{DIM}}(\phi^\star) \mathbf{x} = -\mathbf{x}^T D \mathbf{x} - \mathbf{x}^T W \mathbf{x}. \label{eq:quad2}
\end{equation}
\end{subequations}
Since \( D \) is  a diagonal matrix, the first term on the RHS of Eq.~\eqref{eq:quad2} can be expressed as,

\begin{subequations}
\begin{equation}
\mathbf{x}^T D \mathbf{x} = \sum_{i} d_i x_i^2, \label{eq:quadnew2}
\end{equation}
while the second term containing the weight matrix \( W \) can be expressed as,

\begin{equation}
\mathbf{x}^T W \mathbf{x} = \sum_{i,j} W_{ij} x_i x_j.
\end{equation}
\end{subequations}
Thus, Eq.~\eqref{eq:quad2} can be expressed as, 

\begin{subequations}
\begin{equation}
\mathbf{x}^T D_{DIM}(\phi^\star) \mathbf{x} = -\sum_{i} d_i x_i^2 - \sum_{i,j} W_{ij} x_i x_j. \label{eq:quadnew3}
\end{equation}
Since \( D_{ii} = \sum_{j} W_{ij} \), Eq.~\eqref{eq:quadnew3} can be expressed as, 

\begin{equation}
\mathbf{x}^T D_{DIM}(\phi^\star) \mathbf{x} = -\left( \frac{1}{2} \sum_{i,j} W_{ij} (x_i^2 + x_j^2) + \sum_{i,j} W_{ij} x_i x_j \right). \label{eq:quadnew4}
\end{equation}
Rearranging terms, 

\begin{equation}
\mathbf{x}^T D_{DIM}(\phi^\star) \mathbf{x} = -\frac{1}{2} \sum_{i,j} W_{ij} (x_i^2 + 2x_i x_j + x_j^2).\label{eq:quadnew4}
\end{equation}
Since $(x_i + x_j)^2 = x_i^2 + 2x_i x_j + x_j^2,$
{Eq.~\eqref{eq:quadnew4}} can be expressed as,

\begin{equation}
\mathbf{x}^T D_{\text{DIM}}(\phi^\star) \mathbf{x} = -\frac{1}{2} \sum_{i,j} W_{ij} (x_i + x_j)^2. \label{eq:quod3}
\end{equation}
\end{subequations}
Since \( W_{ij} \geq 0 \), and \( (x_i + x_j)^2 \geq 0\), each term in the summation, and thus, the summation as a whole, is non-negative. Consequently, it can be concluded that Eq.~\eqref{eq:quod3} satisfies the following relationship,

\[
\mathbf{x}^T D_{DIM}(\phi^\star) \mathbf{x} \leq 0, \quad \forall \mathbf{x}.
\]
Thus, \( D_{DIM}(\phi^\star) \) is negative semi-definite. We also note that since $K$ is a positive constant,  $K D_{DIM}(\phi^\star)$ at $\phi^\star \in \left\{\frac{\pi}{2}\right\}$ is also negative semi-definite.

\label{appendix:negative-semi}

\section{\MakeUppercase{Computing Bifurcation Threshold}}
\addcontentsline{toc}{section}{Appendix: Computing Bifurcation Threshold} 

To compute the bifurcation threshold and the corresponding $K_{s,E}$, we compute the absolute value of the phase deviation from $\phi=\frac{\pi}{2}$ i.e., $\Delta = |\phi-\frac{\pi}{2}|$. As expected, $\Delta$ exhibits a rapid increase from 0 to $0.5\pi$ acorss the bifurcation. For the results presented in this work, we observe that setting a threshold of $\Delta = 0.006$ to calculate $K_{s,E}$ provides the best results.  Fig.~\ref{fig:Ks_Critical_zoom} the evolution of $\Delta$ for the example considered in Fig.~\ref{fig:comparison}.

\FloatBarrier
\begin{figure}[htbp!]
    \centering
    \includegraphics[width=0.5\textwidth]{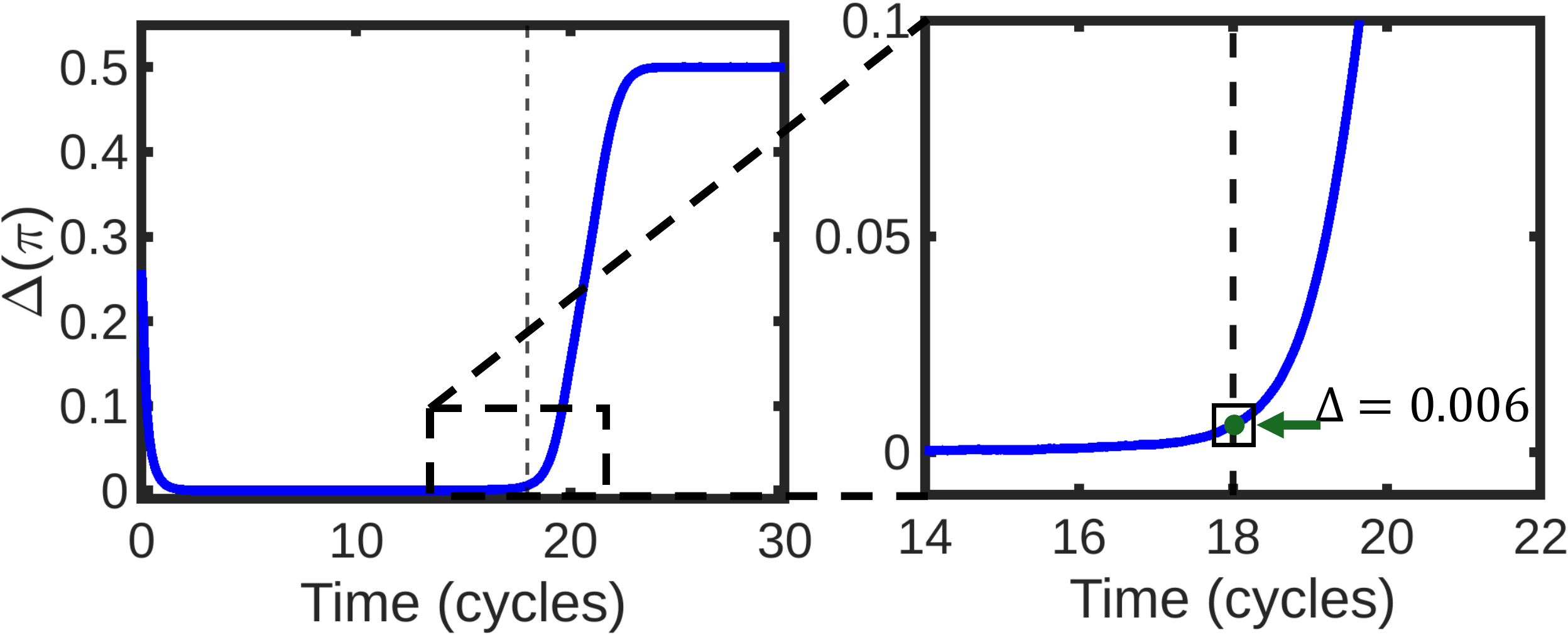}
    \captionsetup{justification=justified} 
    \caption{\justifying (a) Temporal evolution of $\Delta$ for the DIM dynamics presented in Fig.\hyperref[fig:comparison]{~\ref*{fig:comparison}(d)}; (b) Magnified version clearly showing the onset of bifurcation at $\Delta=0.006$.}
    \label{fig:Ks_Critical_zoom}
\end{figure}
\FloatBarrier
\label{appendix:kscritical}

\def\bibsection{\section*{References}}
\bibliography{apssamp}  

\end{document}